\documentclass[runningheads]{llncs}
\usepackage{graphicx}
\usepackage{array}
\usepackage{booktabs}
\usepackage{multirow}
\usepackage{amsmath}
\usepackage{hyperref}
\usepackage{cleveref}
\usepackage{url}

\usepackage{xspace}

\newcommand{\tabref}[2][]{Table#1~\ref{#2}\xspace}
\newcommand{\secref}[2][]{Section#1~\ref{#2}\xspace}

\newcommand{\eqnref}[2][]{Equation#1~\ref{#2}\xspace}

\begin{document}
%

\title{Evaluating a Multi-sense Definition Generation Model for
  Multiple Languages}

\titlerunning{Evaluating a Multi-sense Definition Generation Model}

%
%

\author{Arman Kabiri \and Paul Cook}

\authorrunning{A. Kabiri and P. Cook}

%

\institute{Faculty of Computer Science, University of New Brunswick\\
Fredericton, NB E3B 5A3 Canada\\
\email{\{arman.kabiri,paul.cook\}@unb.ca}}

\maketitle              
\begin{abstract}

Most prior work on definition modelling has not accounted for
polysemy, or has done so by considering definition modelling for a
target word in a given context. In contrast, in this study, we propose
a context-agnostic approach to definition modelling, based on
multi-sense word embeddings, that is capable of generating multiple
definitions for a target word. In further contrast to most prior work,
which has primarily focused on English, we evaluate our proposed
approach on fifteen different datasets covering nine languages from
several language families. To evaluate our approach we consider
several variations of BLEU. Our results demonstrate that our proposed
multi-sense model outperforms a single-sense model on all fifteen
datasets.

\end{abstract}

\section{Introduction}
\label{intro}


The advent of pre-trained distributed word representations, such as
\cite{mikolov2013distributed}, led to improvements in a wide range of
natural language processing (NLP) tasks. One limitation of such word
embeddings, however, is that they conflate all of a word's senses into
a single vector. Subsequent work has considered approaches to learn
multi-sense embeddings, in which a word is represented by multiple
vectors, each corresponding to a sense
\cite{adagram,lee2017muse}. More recent work has considered
contextualized word embeddings, such as \cite{devlin2018bert}, which
provide a representation of the meaning of a word in a given context.

Definition modelling, recently introduced by
\cite{noraset2017definition}, is a specific type of language modelling
which aims to generate dictionary-style definitions for a given
word. Definition modelling can provide a transparent interpretation of
the information represented in word embeddings, and has the potential
to be applied to generate definitions for newly-emerged words that are
not yet recorded in dictionaries.

The approach to definition modelling of \cite{noraset2017definition}
is based on a recurrent neural network (RNN) language model, which is
conditioned on a word embedding for the target word to be defined,
specifically pre-trained word2vec \cite{mikolov2013distributed}
embeddings.  As such, this model does not account for polysemy. To
address this limitation, a number of studies have proposed
context-aware definition generation models
\cite{ni2017learning,gadetsky2018conditional,ishiwatari2018learning,mickus-etal-2019-mark,chang2019does}. In
all of these approaches, the models generate a definition
corresponding to the usage of a given target word in a given context.


In contrast, in this paper we propose a context-agnostic multi-sense
definition generation model. Given a target word type (i.e., without
its usage in a specific context) the proposed model generates multiple
definitions corresponding to different senses of that word. Our
proposed model is an extension of \cite{noraset2017definition} that
incorporates pre-trained multi-sense embeddings. As such, the
definitions that are generated are based on the senses learned by the
embedding model on a background corpus, and reflect the usage of words
in that corpus. Under this setup --- i.e., generating multiple
definitions for each word corresponding to senses present in a corpus
--- the proposed definition generation model has the potential to
generate partial dictionary entries. In order to train the proposed
model, pre-trained sense vectors for a word need to be matched to
reference definitions for that word. We consider two approaches to
this matching based on cosine similarity between sense vectors and
reference definitions.

Recently, \cite{zhu2019multi} propose a multi-sense model for
generating definitions for the various senses of a target word. This
model utilizes word embeddings and coarse-grained atom embeddings to
represent senses \cite{arora2018linearTRIM}, in which atoms are shared
across words. In contrast, we only rely on fine-grained multi-sense
embeddings. To match sense vectors to reference definitions during
training, \cite{zhu2019multi} propose a neural approach, and also
consider a heuristic-based approach that incorporates cosine
similarity between senses and definitions. Our proposed approach to
this matching is similar to their heuristic-based approach, although
we explore two variations of this method. Furthermore,
\cite{zhu2019multi} only consider English for evaluation, whereas we
consider fifteen datasets covering nine languages.


Following \cite{zhu2019multi} we evaluate our proposed model using
variations of BLEU \cite{papineni2002bleu}. We evaluate our model on
fifteen datasets covering nine languages from several families. Our
experimental results show that, for every language and dataset
considered, our proposed approach outperforms the benchmark approach
of \cite{noraset2017definition} which does not model polysemy.

\section{Proposed Model}
\label{section:model}




Here we briefly describe the model of \cite{noraset2017definition},
referred to as the base model, and then present our proposed
multi-sense model which builds on the base model.




The base model is an RNN-based language model which, given a target
word to be defined ($w^*$), predicts the target word's definition ($D
= [w_1, ..., w_T]$). The probability of the \textit{t}th word of the
definition sequence, $w_t$, is calculated based on the previous words
in the definition as well as the word being defined, as shown in
\eqnref{eqn:1}.
\begin{equation}
\label{eqn:1}
  P(D|w^*) = \prod_{t=1}^{T} p(w_t|w_1,...,w_t-1,w^*)
\end{equation}
\noindent
The probability distribution is estimated by a softmax function. The
model further incorporates a character-level CNN to capture knowledge
of affixes. A full explanation of this model is in
\cite{noraset2017definition}.



In the base model, the target word being defined ($w^*$) is
represented by its word2vec word embedding. This reliance on
single-sense embeddings limits the model's ability to generate
definitions for different senses of polysemous target words.  To
address this limitation, we propose to extend the base model by
incorporating multi-sense embeddings, in which each word is
represented by multiple vectors which correspond to different meanings
or senses for that word. Specifically, we replace $w^*$ in
\eqnref{eqn:1} by a sense of the target word, represented as a sense
vector.


Most prior work on definition modelling has considered polysemy
through context-aware approaches
\cite{ni2017learning,gadetsky2018conditional,ishiwatari2018learning,mickus-etal-2019-mark,chang2019does}
that require an example of the target word in context for definition
generation. In contrast, the model we propose is context agnostic (as
is the base model) and is able to generate multiple definitions for a
target word without requiring that specific contexts of the target
word be given in order to generate definitions.



The base model is trained on instances consisting of pairs of a word
--- represented by a word2vec embedding --- and one of its
definitions, i.e., from a dictionary. Our proposed approach is trained
on pairs of a word sense --- represented as a sense vector --- and one
of the corresponding word's definitions. In order to train our
proposed approach, we require a way to associate pre-trained sense
vectors with dictionary definitions, where the number of sense vectors
and definitions is often different for a given word.


We consider two approaches to associating sense vectors with
definitions: definition-to-sense and sense-to-definition. For both
approaches we require a representation of definitions. We represent a
definition as the average of its word embeddings, after removing
stopwords. For each word in the training data, we then calculate the
pairwise cosine similarity between its sense vectors and
definitions. For definition-to-sense, each definition is associated
with the most similar sense vector for the corresponding word. For
sense-to-definition, on the other hand, each sense is associated with
the most similar definition. For both approaches, the selected
sense--definition pairs form the training data.

These approaches to pairing senses and definitions are only used to
create training instances. At test time, to generate definitions for a
given target word, each sense vector for the target word is fed to
the definition generation model, which then generates one definition
for each of the target word's sense vectors.

\section{Materials and Methods}

In this section, we describe the datasets, word and sense embeddings,
and evaluation metrics used in our experiments.

\subsection{Datasets}

In this work, we conduct a multi-lingual study of definition
modelling. We extract monolingual dictionaries for nine languages
covering several language families, from three different sources:
Wiktionary,\footnote{\url{https://en.wiktionary.org}}
OmegaWiki,\footnote{\url{http://www.omegawiki.org}} and WordNet
\cite{miller1998wordnet}.

Wiktionary is a free collaboratively-constructed online dictionary for
many languages. The structure of Wiktionary pages is not consistent
across languages. Extracting word--definitions pairs from Wiktionary
pages for a given language requires a carefully-designed
language-specific parser, which moreover requires some knowledge of
that language to build. We therefore use publicly-available Wiktionary
parsers. We use WikiParsec for English, French, and
German,\footnote{\url{https://github.com/LuminosoInsight/wikiparsec}}
and Wikokit for
Russian,\footnote{\url{https://github.com/componavt/wikokit}} to
extract word--definition pairs for these languages.




OmegaWiki, like Wiktionary, is a free collaborative multilingual
dictionary. In OmegaWiki data is stored in a relational database, and
so language-specific parsers are not required to automatically extract
words and definitions. We extract the word--definition pairs from
OmegaWiki for English, Dutch, French, German, Italian, and Spanish ---
the six languages with the largest vocabulary size in OmegaWiki ---
using the BabelNet Java API
\cite{navigli2012babelnet}.

Finally, we consider WordNets. We only use WordNets for which the
words and definitions are in the same language. We again use the
BabelNet Java API to extract the word--definition entries from English
\cite{miller1998wordnet}, Italian \cite{artale1997wordnet}, and
Spanish \cite{spanish-wordnet} WordNets. We separately extract
word--definition pairs from Greek \cite{stamou2004exploring} and
Japanese \cite{isahara2008development} WordNets.



Properties of the extracted datasets are shown in
\tabref{table-datasets-properties}. Each dataset is partitioned into
train (80\%), dev (10\%), and test (10\%) sets. We ensure that, for
each word in each dataset, all of its definitions are included in only
one of the train, dev, or test sets, so that models are only evaluated
on words that were not seen during training.

\begingroup
\setlength{\tabcolsep}{6pt} 
\renewcommand{\arraystretch}{1} 
\begin{table}
\begin{center}
\caption{\label{table-datasets-properties} The number of words, and
  proportion of polysemous words (PPW) in each dataset.}
\begin{tabular}{lcccccc}
\toprule
\multirow{2}{*}{Language} & \multicolumn{2}{c}{Omega} & \multicolumn{2}{c}{Wiktionary} & \multicolumn{2}{c}{WordNet}\\
 & \#Words & PPW & \#Words & PPW & \#Words & PPW\\
\midrule
Dutch & 13093 & 0.18 & -- & -- & -- & --\\
English & 17000 & 0.20 & 17000 & 0.27 & 20000 & 0.18\\
French & 15869 & 0.17 & 20000 & 0.26 & -- & --\\
German & 13338 & 0.12 & 16000 & 0.26 & -- & --\\
Greek & -- & -- & -- & -- & 11517 & 0.26\\
Italian & 18351 & 0.21 & -- & -- & 16290 & 0.22\\
Japanese & -- & -- & -- & -- & 20000 & 0.30\\
Russian & -- & -- & 15000 & 0.17 & -- & --\\
Spanish & 17000 & 0.19 & -- & -- & 18934 & 0.12\\
\bottomrule
\end{tabular}
\end{center}
\end{table}
\endgroup


\subsection{Word and Sense Embeddings}

Following \cite{noraset2017definition}, we use word2vec embeddings in
the singe-sense definition generation model (i.e., the base
model). For the proposed multi-sense models, we utilize AdaGram
embeddings \cite{adagram}. AdaGram is a non-parametric Bayesian
extension of Skip-gram which learns a variable number of sense vectors
for each word, unlike many multi-sense embedding models which learn a
fixed number of senses for every word. Note that although here we use
AdaGram, any multi-sense embedding method could potentially be
used.\footnote{In preliminary experiments with MUSE embeddings
  \cite{lee2017muse} we found MUSE to perform poorly compared to
  AdaGram, and so only report results for AdaGram here.}


For each language, word2vec and AdaGram embeddings are trained on the
most recent Wikipedia dumps as of January
2020.\footnote{\url{https://dumps.wikimedia.org}} We extract plain
text from these dumps,
and then pre-process and tokenize the corpora using tools from
AdaGram,\footnote{\url{https://github.com/sbos/AdaGram.jl/blob/master/utils/tokenize.sh}}
modified for multilingual support, except in the case of Japanese
where we use the Mecab
tokenizer.\footnote{\url{https://github.com/jordwest/mecab-docs-en}}
The resulting corpora range in size from roughly 86 million tokens for
Greek to 3.7 billion tokens for English. The same pre-processing and
tokenization is also applied to the datasets of words and definitions
extracted from dictionaries.


We train word2vec embeddings using Gensim with its default
parameters.\footnote{\url{https://radimrehurek.com/gensim/}} We also
use the default parameter settings for AdaGram. To obtain
representations for words, as opposed to senses, from AdaGram sense
embeddings, as required to form representations for definitions
(\secref{section:model}), we take the most frequent sense vector of
each word (as indicated by Adagram) as the representation of the word
itself.




\subsection{Evaluation Metrics}

BLEU \cite{papineni2002bleu} has been widely used for evaluation in
prior work on definition modelling
\cite{noraset2017definition,ishiwatari2018learning,ni2017learning}.
BLEU is a precision-based metric that measures the overlap of a
generated sequence (here a definition) with respect to one or more
references. For multi-sense models, we calculate BLEU as the average
BLEU score over each generated definition.


While BLEU is appropriate for evaluation of single-sense definition
generation models, it does not capture the ability of a model to
produce multiple definitions corresponding to different senses of a
polysemous word. We therefore also consider a recall-based variation
of BLEU, known as rBLEU, in which the generated and reference
definitions are swapped \cite{zhu2019multi}, i.e., the overlap of a
reference definition is measured with respect to the generated
definition(s). For each target word, we calculate rBLEU as the average
rBLEU score for each of its reference definitions (for both single and
multi-sense models).





In addition to precision-based BLEU, and recall-based rBLEU, we report
the harmonic mean of BLEU and rBLEU, referred to as fBLEU.

\section{Results}

In this section, we present experimental results comparing the
proposed multi-sense definition generation models against the
single-sense base model \cite{noraset2017definition}. All models are
trained using parameter settings from \cite{noraset2017definition},
i.e., a two-layer LSTM as the RNN component with 300 units in each
level; a character-level CNN with kernels of length 2--6 and size
$\{10, 30, 40, 40, 40\}$ with a stride of 1; and Adam optimization
with a learning rate of 0.001.

\begingroup
\setlength{\tabcolsep}{2.2pt} 
\renewcommand{\arraystretch}{1} 
\begin{table}
\begin{center}
\caption{\label{table-results} BLEU, rBLEU, and fBLEU for the
  single-sense definition generation model (base) and the proposed
  multi-sense models using sense-to-definition (S2D) and
  definition-to-sense (D2S) for each dataset. The best result for each
  evaluation metric and dataset is shown in boldface.}
\begin{tabular}{ll|ccc|ccc|ccc}
\toprule

\multirow{2}{*}{Lang.} & \multirow{2}{*}{Model} & \multicolumn{3}{c|}{OmegaWiki} & \multicolumn{3}{c|}{Wiktionary} & \multicolumn{3}{c}{WordNet}\\

 &  & BLEU & rBLEU & fBLEU & BLEU & rBLEU & fBLEU & BLEU & rBLEU & fBLEU\\

\hline

\multirow{3}{*}{DE}
& base & 12.12 & 11.55 & 11.83 & 11.35 & 08.80 & 09.91 & -- & -- & --\\
& S2D & 12.43 & 16.26 & 14.09 & \textbf{15.00} & 15.82 & \textbf{15.40} & -- & -- & --\\
& D2S & \textbf{12.44} & \textbf{16.83} & \textbf{14.31} & 14.07 & \textbf{16.54} & 15.21 & -- & -- & --\\

\hline

\multirow{3}{*}{EL}
& base & -- & -- & -- & -- & -- & -- & \textbf{13.21} & 12.06 & 12.61\\
& S2D & -- & -- & -- & -- & -- & -- & 12.44 & 12.85 & 12.64\\
& D2S & -- & -- & -- & -- & -- & -- & 13.08 & \textbf{13.63} & \textbf{13.35}\\

\hline

\multirow{3}{*}{EN}
& base & 14.74 & 14.32 & 14.53 & 20.21 & 16.88 & 18.40 & 13.78 & 12.77 & 13.26\\
& S2D & 14.23 & 16.02 & 15.07 & 18.88 & 16.99 & 17.89 & 12.85 & 13.09 & 12.97\\
& D2S & \textbf{15.22} & \textbf{17.80} & \textbf{16.41} & \textbf{21.49} & \textbf{19.78} & \textbf{20.60} & \textbf{13.84} & \textbf{14.84} & \textbf{14.32}\\

\hline

\multirow{3}{*}{ES}
& base & \textbf{17.68} & 17.70 & 17.69 & -- & -- & -- & \textbf{26.46} & 24.69 & 25.54\\
& S2D & 16.52 & 19.00 & 17.67 & -- & -- & -- & 25.80 & \textbf{28.14} & \textbf{26.92}\\
& D2S & 17.54 & \textbf{20.28} & \textbf{18.81} & -- & -- & -- & 25.68 & 27.97 & 26.78\\

\hline

\multirow{3}{*}{FR}
& base & \textbf{12.58} & 12.66 & 12.62 & 63.48 & 59.87 & 61.62 & -- & -- & --\\
& S2D & 11.70 & 14.26 & 12.85 & 63.56 & 60.00 & 61.73 & -- & -- & --\\
& D2S & 11.94 & \textbf{14.82} & \textbf{13.23} & \textbf{64.12} & \textbf{60.41} & \textbf{62.21} & -- & -- & --\\

\hline

\multirow{3}{*}{IT}
& base & \textbf{12.29} & 11.93 & 12.11 & -- & -- & -- & 21.33 & 20.65 & 20.98\\
& S2D & 11.43 & 13.61 & 12.43 & -- & -- & -- & 20.35 & 23.67 & 21.88\\
& D2S & 11.74 & \textbf{13.95} & \textbf{12.75} & -- & -- & -- & \textbf{21.96} & \textbf{25.10} & \textbf{23.43}\\

\hline

\multirow{3}{*}{JA}
& base & -- & -- & -- & -- & -- & -- & 10.13 & 08.50 & 09.24\\
& S2D & -- & -- & -- & -- & -- & -- & \textbf{11.53} & \textbf{11.96} & \textbf{11.74}\\
& D2S & -- & -- & -- & -- & -- & -- & 09.42 & 09.37 & 09.39\\

\hline

\multirow{3}{*}{NL}
& base & 14.37 & 14.04 & 14.20 & -- & -- & -- & -- & -- & --\\
& S2D & 13.49 & 15.88 & 14.59 & -- & -- & -- & -- & -- & --\\
& D2S & \textbf{14.46} & \textbf{17.07} & \textbf{15.66} & -- & -- & -- & -- & -- & --\\

\hline

\multirow{3}{*}{RU}
& base & -- & -- & -- & 47.04 & 46.04 & 46.53 & -- & -- & --\\
& S2D & -- & -- & -- & 46.24 & 46.69 & 46.46 & -- & -- & --\\
& D2S & -- & -- & -- & \textbf{47.52} & \textbf{48.09} & \textbf{47.80} & -- & -- & --\\

\bottomrule

\end{tabular}
\end{center}
\end{table}
\endgroup

To generate definitions at test time, for each word and sense for the
single-sense and multi-sense models, respectively, we sample tokens at
each time step from the predicted probability distribution with a
temperature of 0.1. We compute BLEU, rBLEU, and fBLEU for each word,
and then the average of these measures over all words in a dataset. We
repeat this process 10 times, and report the average scores over these
10 runs.






Results are shown in \tabref{table-results}. Focusing on fBLEU, for
every dataset, the best results are obtained using a multi-sense model
--- i.e., sense-to-definition (S2D), or definition-to-sense
(D2S). Moreover, for every dataset, D2S improves over the base model.
These results show that definition modelling can be improved by
accounting for polysemy through the incorporation of multi-sense
embeddings.

To qualitatively compare the base model and the proposed model, we
consider the definitions generated for the word \emph{state}. The
following three definitions are generated for this word by the base
model: (1) \emph{a state of a government}, (2) \emph{to make a certain
  or permanent power}, and (3) \emph{to make a certain or
  administrative power}. In contrast, the proposed multi-sense model
using D2S generates the following three definitions, which appear to
capture a wider range of the usages of the word \emph{state}: (1)
\emph{a place of government}, (2) \emph{a particular region of a
  country}, and (3) \emph{a particular place of time}.





Comparing S2D and D2S in terms of fBLEU, we observe that D2S often
performs better. The number of sense vectors learned by Adagram for a
given word is on average higher than the number of reference
definitions available for that word, for every dataset. We hypothesize
that the poor performance of S2D relative to D2S could therefore be
due to sense vectors being associated with inappropriate definitions.





rBLEU is a recall-based evaluation metric that indicates the extent to
which the reference definitions are covered by the generated
definitions. A multi-sense definition generation model --- which
produces multiple definitions for a target word --- is therefore
particularly advantaged compared to a single-sense model --- such as
the base model --- which produces only one, with respect to this
metric. Indeed, we see that for every dataset, both
S2D and D2S, outperform the base model
in terms of rBLEU. BLEU, on the other hand, is a precision-based
metric that indicates whether a generated definition contains material
present in the reference definitions. The improvements of the
multi-sense models over the base model with respect to rBLEU do not
substantially impact BLEU --- as observed by the overall higher fBLEU
obtained by the multi-sense models. Overall, these results indicate
that a multi-sense model is able to generate definitions that better
reflect the various senses of polysemous words than a single-sense
model, without substantially impacting the quality of the individual
generated definitions.




\section{Conclusions}

Definition modelling is a recently-introduced language modelling task
in which the aim is to generate dictionary-style definitions for a
given word. In this paper, we proposed a multi-sense context-agnostic
definition generation model which employed multi-sense embeddings to
generate multiple senses for polysemous words. In contrast to most
prior work on definition modelling which focuses on English, we
conducted a multi-lingual study including nine languages from several
language families. Our experimental results demonstrate that our
proposed multi-sense model outperforms a single-sense baseline
model. Code and datasets for these experiments is
available.\footnote{\url{https://github.com/ArmanKabiri/Multi-sense-Multi-lingual-Definition-Modeling}}
In future work, we intend to consider incorporating alternative
approaches to learning multi-sense embeddings into our model, as well
as alternative approaches to associating sense vectors to definitions
for constructing training instances.






%

\bibliographystyle{splncs04}
\bibliography{biblography.bib}






\end{document}